\documentclass{bmvc2k}

\title{Multi-Task Edge Prediction in Temporally-Dynamic Video Graphs}

\addauthor{Osman Ülger}{o.ulger@uva.nl}{1}
\addauthor{Julian Wiederer}{julian.wiederer@mercedes-benz.com}{2}
\addauthor{Mohsen Ghafoorian}{mohsenghafoorian@gmail.com}{}
\addauthor{Vasileios Belagiannis}{vasileios.belagiannis@fau.de}{3}
\addauthor{Pascal Mettes}{p.s.m.mettes@uva.nl}{1}

\addinstitution{University of Amsterdam\\
 Amsterdam, NL
}
\addinstitution{Mercedes-Benz\\
    Stuttgart, DE
}
\addinstitution{Friedrich-Alexander-Universität Erlangen-Nürnberg,\\
    Erlangen, DE
}

\runninghead{Ulger et al.}{Multi-Task Edge Prediction in Temporally-Dynamic Graphs}

\def\eg{\emph{e.g}\bmvaOneDot}
\def\ie{\emph{i.e.}\bmvaOneDot}

\def\etal{\emph{et al}\bmvaOneDot}

\usepackage{booktabs}
\usepackage{comment}
\usepackage{times}
\usepackage{epsfig}
\usepackage{graphicx}
\usepackage{amsmath}
\usepackage{amssymb}
\usepackage{kbordermatrix} 
\usepackage{xcolor,colortbl}
\usepackage{xspace}
\usepackage{dsfont}     
\usepackage{comment}
\definecolor{mygreen}{rgb}{0.3, 0.9, 0.4}
\usepackage{multirow}
\usepackage{wrapfig,lipsum,booktabs}

\usepackage[capbesideposition=outside,capbesidesep=quad]{floatrow}

\definecolor{Gray}{gray}{0.9}

\newcommand{\boldparagraph}[1]{\vspace{0.1em}\noindent{\bf #1}}
\newcommand{\Ind}[1]{\mathds{1}\!\left[#1\right]}

\begin{document}

\maketitle

\begin{abstract}
Graph neural networks have shown to learn effective node representations, enabling node-, link-, and graph-level inference. Conventional graph networks assume static relations between nodes, while relations between entities in a video often evolve over time, with nodes entering and exiting dynamically. In such temporally-dynamic graphs, a core problem is inferring the future state of spatio-temporal edges, which can constitute multiple types of relations. To address this problem, we propose MTD-GNN, a graph network for predicting temporally-dynamic edges for multiple types of relations. We propose a factorized spatio-temporal graph attention layer to learn dynamic node representations and present a multi-task edge prediction loss that models multiple relations simultaneously. The proposed architecture operates on top of scene graphs that we obtain from videos through object detection and spatio-temporal linking. Experimental evaluations on ActionGenome and CLEVRER show that modeling multiple relations in our temporally-dynamic graph network can be mutually beneficial, outperforming existing static and spatio-temporal graph neural networks, as well as state-of-the-art predicate classification methods. Code is available at \href{https://github.com/ozzyou/MTD-GNN}{https://github.com/ozzyou/MTD-GNN}.
\end{abstract}

\section{Introduction}
\label{sec:intro}
Graph neural networks (GNN) have become an established framework for visual recognition and understanding, with applications such as action recognition \cite{situationrec, wang2018videos, stgcnaction, actionrec}, semantic segmentation \cite{gnnsegmentation, gnnseg2}, and visual relation detection \cite{visualrelationship}. A common assumption in GNNs is that nodes are stationary or at least always present. In practice, however, especially in the video domain, visual relations evolve dynamically over time. Entities, modeled as graph nodes, can enter or exit scenes, while edges have evolving semantics. In this paper, we address the problem of predicting edge labels in dynamically-evolving spatio-temporal graphs. 

\begin{figure}[t]
\centering
\includegraphics[width=0.8\linewidth]{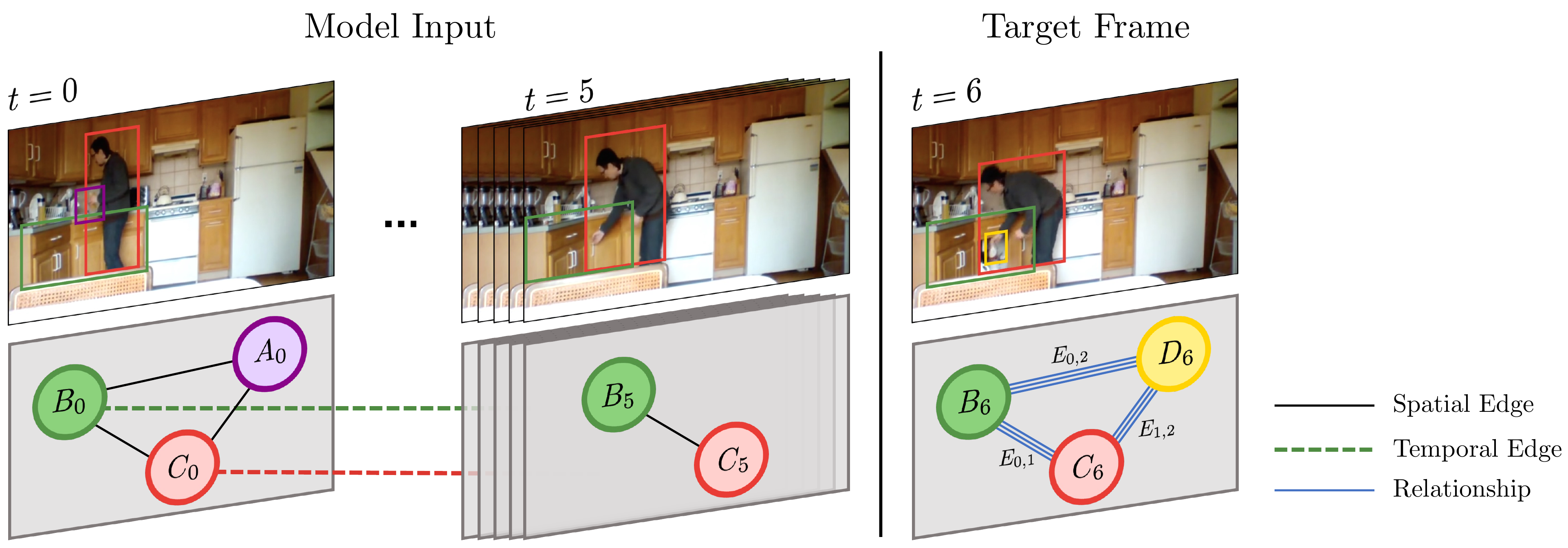}\\
\caption{\textbf{Problem setting.} The model input is a spatio-temporal graph $\mathcal{G}_{\text{input}}$, built from object detections in multiple, adjacent frames. Visual feature representations of detections are modeled as graph nodes which can enter and exit scenes over time. Our approach is able to handle such dynamic graph changes. We predict relationships between objects in the target frame, which follows the last input frame. Based on the input, our model is able to predict multiple relation types - pictured are three - depicted with $E_{i,j}$ for objects $i$ and $j$.}
\label{fig:approach_overview}
\end{figure}

Multiple works have investigated spatio-temporal graph neural networks dealing with the temporal dynamics of graphs, for example for skeleton-based action recognition~\cite{li2019spatio, stgcnaction} and object-action relations in video~\cite{humanobjectinteraction2, humanobjectinteraction}. While such networks incorporate temporal dynamics, there is typically one set of entities that remains present in the scene across time. In contrast, we are interested in a more challenging setting where entities may enter and/or exit the scene over time. Such a setting is relevant for real-world applications such as autonomous driving~\cite{ownermemberrel, roaddataset}. Fig.~\ref{fig:approach_overview} illustrates our setup. We propose a graph network that does not rely on stationary node assumptions and enables learning multiple relations simultaneously.

In this work, we introduce the task of future state multi-relational edge label prediction in temporally-dynamic graphs. Moreover, we introduce a Multi-task Temporally-Dynamic Graph Neural Network (MTD-GNN), a graph network centered around a factorized spatio-temporal graph attention layer as a natural solution to learn with dynamic node sets in both space and time, inspired by static graph attention~\cite{GAT}. On top of the graph attention layers, MTD-GNN learns multiple relation types as a weighted multi-task optimization. The graph network is learned on top of spatio-temporal interaction graphs constructed through detection and temporal linking. Experiments on CLEVRER~\cite{clevrer} and Action Genome~\cite{ji2020action} demonstrate that our approach can handle temporally-dynamic spatio-temporal scene graphs, while also outperforming most existing static, as well as state-of-the-art methods. Additionally, we find that learning in a multi-task manner can boost the model's performance on individual tasks.
\section{Related Work}
\boldparagraph{Static graph networks.}
Graph neural networks denote a family of representation learning algorithms on graph-structured data. Graph networks learn node-level representations while abiding by the permutation invariant nature of graphs. Well-known instantiations of graph neural networks include the original Graph Neural Network by Scarselli~\etal~\cite{GNN}, Graph Convolutional Networks~\cite{gcn}, and Graph Attention Networks~\cite{GAT}. Commonly, each node aggregates information from its neighbours within a graph layer, accompanied by a shared weight matrix. By stacking multiple graph layers, node representations are learned using information from nodes throughout the graph. Given learned node representations from stacked graph layers, inference can be performed on nodes~\cite{Gong_2019_CVPR, zhao2019semantic, zhang_2019_heterogeneous}, links~\cite{bordes, hu2019neural, entityembedding, zhang2017weisfeiler, zhang2018link}, and graphs~\cite{chen2019multi, messagepassing, lee_2019_settransformer}. Close to our approach are works on link prediction~\cite{Adamic_friends_2003, kim_correlation_clustering_2011} in graph neural networks~\cite{Nickel_2016}, \eg for recommendation~\cite{berg2018graph, socialrecommendation, koren2009matrix}. For example, Wang~\etal~\cite{wang_kgat_2019} use an attention mechanism in user-item graphs to predict user recommendations as links. A number of works have also investigated multiple relation types between nodes in graph networks \cite{Gong_2019_CVPR, fashion}. While these works target static multi-relational learning, we focus on multi-relational learning in temporally-dynamic graphs. The work of Kipf~\etal~\cite{kipf2018neural} uses edges between objects to model their latent interactions as a useful encoding to predict object dynamics. This example shows the expressive power of edges for downstream tasks. Closest to our work, Kim~\etal~\cite{kim2019edge} follow a more explicit formulation of edges. They use an edge-labeling graph neural network with edges representing assignments to clusters in supervised and semi-supervised image classification. The mentioned methods are designed for static graphs and therefore rely on a constant number of input nodes. Similarly, we seek to label edges. However, rather than inferring on edges in static graphs, we do so in temporally-dynamic spatio-temporal graphs, where nodes and relations evolve over time.

\boldparagraph{Spatio-temporal graph networks.}
Multiple works have investigated extensions of graph networks to the spatio-temporal domain, for tasks such as activity recognition \cite{herzig2019spatio, li2019spatio, stgcnaction} and traffic forecasting \cite{chen2020multi, diao2019dynamic, trafficforecasting}.
Spatio-temporal graph networks are commonly tackled using recurrent networks or through spatio-temporal convolutions. For example, structural-RNNs are used to learn about interactions between humans and objects in videos \cite{structuralrnn}. Likewise, Wang~\etal~\cite{wang2018videos} model interactions in videos using convolutional appearance features as graph nodes and Yang~\etal~\cite{stgcnidentification} use a spatio-temporal graph convolution network for person re-identification. In this paper, we also focus on spatio-temporal graphs, but consider the more challenging scenario where nodes can enter and exit scenes over time and where nodes exhibit multiple relation types. Different from Xu~\etal~\cite{tgat}, who used functional time encodings for node classification and link prediction tasks, our method operates directly on the graph for multi-relational edge prediction.

\boldparagraph{Scene graph generation.} Scene graphs have many applications, such as in image captioning~\cite{imagecaptioningwang, sgautoencoder, Kim_2019_CVPR}, visual question answering~\cite{relationaware} and image generation~\cite{imagegensg, interactivesg}. To construct a scene graph, classifiers are trained to predict the categories of object detections and their relationships. Different from scene graph generation, we seek to predict the future state of object relationships based on a temporally-dynamic spatio-temporal graph, rather than relationships with a known state in a static scene graph. Nevertheless, we provide comparisons with state-of-the-art predicate classifiers of the scene graph generation task in Sec.~\ref{compev}.
\section{MTD-GNN} \label{method}
\begin{figure*}[t]
  \centering
  \includegraphics[width=1\linewidth]{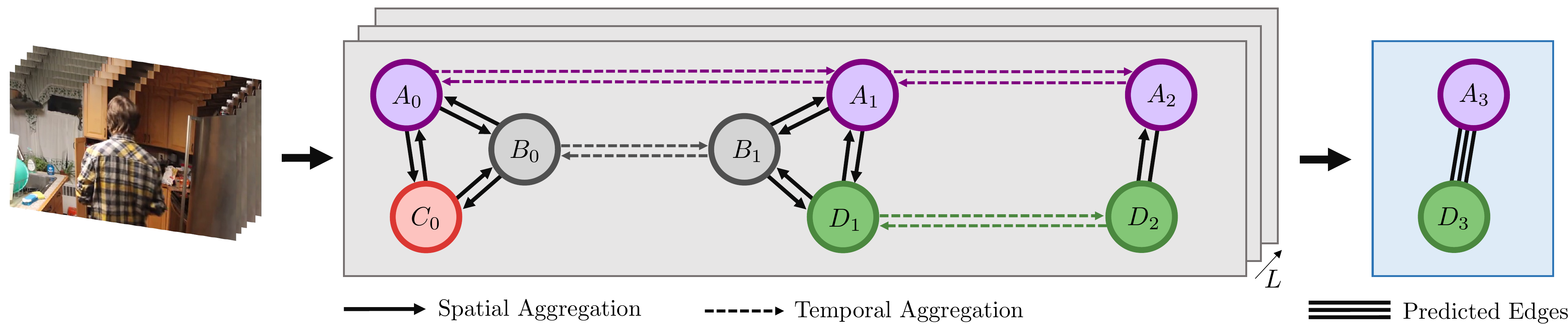}\\
  \caption{
  \textbf{Proposed architecture.} Nodes in the spatio-temporal graph are attended in space and time over $L$ graph attention layers. Each pair of updated nodes is fed to the edge prediction module, which predicts a pair-wise edge value for all relations $R$.}
\label{fig:aggregation}
\end{figure*}

For the problem of temporally-dynamic edge prediction of future states, we construct a spatio-temporal graph $\mathcal{G}_{\text{input}}$ with $F$ timesteps from a video with $T$ frames, denoted as $\mathcal{G}_{\text{input}} = \{G_0, \dots, G_{F-1}\}$, where $F < T$. 
Rather than representing an object with a unique node, we define a new node for each detected object at each timestep. Let $\mathcal{N}$ denote the total number of detected nodes in the graph and $N_i$ the nodes at timestep $i$. We denote the set of spatial edges as $\mathcal{E}^s = \{\mathcal{E}^s_0,...,\mathcal{E}^s_{F-1}\}$ and the temporal edges as $\mathcal{E}^t = \{\mathcal{E}^t_0,...,\mathcal{E}^t_{F-2}\}$. 
Spatial edges are between different objects at the same timestep, temporal edges are between the same object in consecutive timesteps (see Fig.~\ref{fig:aggregation}). 
Our goal is to predict the spatial edge labels between pairs of nodes for all relations $r \in \mathcal{R}$ in the final frame $T$ using $\mathcal{G}_{\text{input}}$. In other words, the model observes object interactions until timestep $F$ and predicts their future state at timestep $T$. We compute a pairwise categorical edge matrix $E_r \in \mathbb{R}^{N_F \times N_F}$, where $N_F$ is the last known number of detected nodes in $\mathcal{G}_{\text{input}}$. In our graph network, we first perform graph attention in space and time simultaneously to learn node-level representations, after which we utilize the node representations to optimize edge prediction for multiple relation types. 

\boldparagraph{Spatio-temporal scene graph generation.}
To operate on video graphs, we first construct a temporally-dynamic spatio-temporal graph $\mathcal{G}_{\text{input}}$ from a video (Fig. \ref{fig:aggregation}). Given pre-trained Mask-RCNN~\cite{MaskRCNN} and Faster-RCNN~\cite{renNIPS15fasterrcnn} backbones for CLEVRER and ActionGenome, respectively, we extract $d$-dimensional feature representations of each detected object such that the total set of node features of $\mathcal{G}_{\text{input}}$ becomes $\textbf{v} = \{\vec{v_1}, \vec{v_2}, \dots, \vec{v_\mathcal{N}}\}$ with $\vec{v}_i \in \mathbb{R}^{d}$. In our experiments, we set $d$ to $256$ and $2048$ for CLEVRER and ActionGenome, respectively.
We then generate a joint spatio-temporal adjacency matrix $\textbf{A} \in \mathbb{R^{\mathcal{N} \times \mathcal{N}}}$ with adjacencies between spatial and temporal neighbours over consecutive timesteps.
Specifically, we connect all detected objects in a particular frame $z$ spatially, while temporally connect each object only with itself if detected in frame $z-1$ or $z+1$. 
We apply Hungarian matching between frames $(\textbf{v}_{z}, \textbf{v}_{z+1})$ and $(\textbf{v}_{z}, \textbf{v}_{z-1})$ to ensure an accurate appearance-based connection without accessing its ground truth location. 
When the object detector misses an object, its relationships with detected nodes are ignored. 
False positive detections can occur when building the spatio-temporal graph, but corresponding predicted edges are not evaluated. 
We ensure this by also applying Hungarian matching between ground truth and proposed bounding boxes. 

\boldparagraph{Factorized spatio-temporal graph attention.}
Using graph attention~\cite{GAT}, we seek to learn node representations that allow for invariance to the number of neighboring nodes.
For each node, we compute hidden representations by attending over its corresponding spatio-temporal neighbours (Fig. \ref{fig:aggregation}).
We propose a factorized multi-headed spatio-temporal graph attention layer that takes as input the set of node features $\textbf{v} \in \mathbb{R}^{\mathcal{N}\times D}$ and outputs features $\textbf{h} = \{\vec{h'_1}, \vec{h'_2}, \dots, \vec{h'_\mathcal{N}}\}$ with $\vec{h}'_i \in \mathbb{R}^{D'}$ averaged over $K$ attention heads and latent dimension $D'$:

\begin{equation}
    \vec{h}_{i}^k = \sum\limits_{j \in A_i} \Ind{A_{ij} \in \mathcal{E}^s} \alpha_{ij}^k \textbf{W}_{j}^k v_{j} \; + \; \Ind{A_{ij} \in \mathcal{E}^t} \gamma_{ij}^k \textbf{W}_{j}^k v_{j} \enspace,
\end{equation}

where $\alpha_{ij}^k$ is the spatial and $\gamma_{ij}^k$ the temporal attention coefficient computed for node pairs $(v_i, v_j)$ in attention head $k$, $\textbf{W}_j^k \in \mathbb{R}^{D' \times D}$ is a weight matrix, 
$\mathcal{E}^s$ and $\mathcal{E}^t$ are mutually exclusive sets of spatial and temporal connections, and $\Ind{C} = \{1\;\text{if}\;C=true,\  0\;\text{else}\}$ is the indicator function. 
The key idea of our factorized spatio-temporal graph attention is the separation between relational \emph{spatial} information as well as \emph{temporal} object information, thus enabling a richer graph layer.

We obtain the final node representation by averaging over the output features of the $K$ attention heads and by applying a sigmoid non-linearity $\sigma$:

\begin{equation}
    \begin{split}
      & \vec{h}'_{i,0} = \sigma\Bigg(\frac{1}{K}\sum\limits_{k=1}^K \vec{h}_{i}^k\Bigg),\quad\quad
      \vec{h}'_{i,L} = \sigma\Bigg(\frac{1}{K}\sum\limits_{k=1}^K \vec{h}_{i,  L-1}^k\Bigg), \quad\quad L \geq 1 \enspace.
    \end{split}
\end{equation}

After one iteration, each node is informed about its first-order neighbors. By performing factorized graph attention repeatedly, information of higher-order neighborhoods is included.

\boldparagraph{Multi-relational edge learning.}
The graph with all detected objects in target frame $T$ is assumed to be fully connected. 
A prediction is made for all pair-wise connections, where edges are undirected and task-dependent. 
We aim to infer multiple types of relations for each edge simultaneously by learning task-specific fully-connected layers for each relation $r \in R$ using the node representation outputs from graph attention layers. 
We obtain a loss value for each separate task $r$ per sample by evaluating the predicted task-specific edges $E_r$ with the respective ground truth labels $Y_r$, resulting in a total loss represented as:

\begin{equation}
  \mathcal{L}_\text{total} = \sum\limits_{r \in R} \sum\limits_{i \in E_r}\sum\limits_{j \in E_{r}^{i}}\mathcal{L}(E_{r}^{ij}, Y_{r}^{ij}) 
  \quad \text{with} \quad 
  E_{r}^{ij}= E_{r}^{ji} = \psi_{r}\Big(\frac{1}{2}(\vec{h}'_{i,L} + \vec{h}'_{j,L}) + b\Big) \enspace,
\end{equation}

where $\vec{h}'_{i,L}$ and $\vec{h}'_{j,L}$ are the final output features of nodes $i$ and $j$, respectively, $b$ is a bias vector and $\psi$ constitutes the fully-connected layers with non-linear activations. For each edge, we ensure permutation invariance for its two nodes by averaging their respective representations. Some datasets contain a large number of objects and therefore have many edges to predict. Depending on the edge type, this can lead to class imbalance, \eg with collision. To balance the losses across different tasks, we outline a \textit{prioritized loss}, which emphasizes the edge class which is less frequent in the training set. The binary cross-entropy (BCE) losses of over-represented class $o$ for an individual task $r$ are down-weighted by normalizing it with the number of ground truth labels of the larger class in the inferred frame, represented as:
\begin{equation}
  \mathcal{L}_\text{prio} = \big(\Ind{Y_{r}^i = 0} o^{-1} + \Ind{Y_{r}^i = 1}\big) \cdot \mathcal{L}_\text{BCE}(E_{r}^{i}, Y_{r}^{i}) \enspace.
\end{equation}
\section{Experiments}
In our experimental evaluation, we perform a series of ablation studies to investigate the core components of our MTD-GNN, a comparative evaluation to existing approaches that generalize to our setting and qualitative analyses showing success and failure cases.

\subsection{Experimental Setup}
\boldparagraph{Datasets.}
We evaluate on the synthetic CLEVRER dataset~\cite{clevrer} and real-world dataset Action Genome~\cite{ji2020action}, as both datasets contain both dynamic object interactions and multiple edge types. 
CLEVRER features multiple object-object relation types, whereas Action Genome focuses on multiple human-object relation types. In both datasets, a target is defined for each pair of objects that share a spatial edge. 
For CLEVRER, we consider two types of prospective, indirected relations: collision (contacting) and relative motion (spatial). 
The goal is to predict the future state of these relations among objects present in a particular scene. 
To achieve this setting, we leave out $X$ frames in the model's input, where $X$ is uniformly sampled between 5 and 20.
For Action Genome, we follow the relation types from the original paper, which constitute a total of 25 human-object directional relationship classes, divided into three \textit{attention}, six \textit{spatial} and seventeen \textit{contacting} relationship types. 
In the spatial and contacting categories, targets can be multi-label. 
For example, an object might be beneath, behind, and on the side of the human at the same time. 
Here, the relationships in the final frame are predicted. Frames with less than two objects are omitted, as their scene graphs lack edges. 

\boldparagraph{Implementation.}
All models and baselines are implemented in Python and PyTorch 1.7 \cite{pytorch}. Adam \cite{adamoptimizer} is used for optimization, with an initial learning rate of $\eta_0 = 0.001$ which is decreased every epoch $e$ following $\eta_e = \eta_{e-1} / (1+0.9e)$. Each model is trained for 100 and 10 epochs for CLEVRER and Action Genome respectively. Due to its imbalanced nature, prioritized loss is used for CLEVRER, while we train with a BCE loss on Action Genome. 

\subsection{Ablation Studies}
\begin{table*}
\centering
\scriptsize
\renewcommand{\arraystretch}{1.0}
\newcommand{\csp}{\hskip 2em}
\floatbox[\capbeside]{table}[0.7\columnwidth]
  {\caption{\textbf{Overview of all ablation studies on CLEVRER~\cite{clevrer}} with the following insights: (i) fewer graph dimensions are beneficial; (ii) five attention heads help to balance complexity and generalization; (iii) one attention layer is all you need; and (iv) multi-task learning is favored over individual optimization.}
\label{tab:ablations}}%
{\begin{tabular}{l ccc@{\csp} ccc}
\toprule
 & \multicolumn{3}{c}{\textbf{Collision prediction}} & \multicolumn{3}{c}{\textbf{Relative motion}}\\
 & F1 $\uparrow$ & AP $\uparrow$ & AUC $\uparrow$ & F1 $\uparrow$ & AP $\uparrow$ & AUC $\uparrow$\\
\rowcolor{Gray}
\multicolumn{7}{l}{\textbf{Latent Dimensions}} \\
256 & \textbf{0.505} & \textbf{0.441} & \textbf{0.668} & \textbf{0.798} & \textbf{0.812} & \textbf{0.672} \\
512 & 0.472 & 0.417 & 0.624 & 0.780 & 0.810 & 0.668 \\
1024 & 0.436 & 0.395 & 0.601 & 0.776 & 0.816 & 0.674 \\
\rowcolor{Gray}
\multicolumn{7}{l}{\textbf{Attention Heads}} \\
3 & 0.458 & 0.373 & 0.589 & \textbf{0.798} & 0.812 & 0.672 \\
5 & \textbf{0.505}  & \textbf{0.441} & \textbf{0.668} & 0.786 & 0.819 & 0.676 \\
7 & 0.426  & 0.446 & 0.642 & 0.784 & 0.824 & \textbf{0.687} \\
9 & 0.440 & 0.435 & 0.642 & 0.766 & \textbf{0.826} & 0.684 \\
\rowcolor{Gray}
\multicolumn{7}{l}{\textbf{Attention Layers}} \\
1 & \textbf{0.505}  & \textbf{0.441} & \textbf{0.668} & \textbf{0.798} & \textbf{0.812} & \textbf{0.672} \\
2 & 0.459 & 0.340 & 0.517 & 0.780 & 0.786 & 0.616 \\
3 & 0.485 & 0.341 & 0.520 & 0.797 & 0.790 & 0.613 \\
\rowcolor{Gray}
\multicolumn{7}{l}{\textbf{Learning Method}} \\
Single-task & 0.505 & 0.441 & 0.668 & 0.798 & 0.812 & 0.672 \\
Multi-task & \textbf{0.594} & \textbf{0.607} & \textbf{0.768} & \textbf{0.839} & \textbf{0.820} & \textbf{0.688} \\
\bottomrule
\end{tabular}}
\end{table*}

\boldparagraph{Attention dimensionality.} 
The attention mechanism is trained with a latent representation of the original input $\vec{h} \in \mathbb{R}^{N \times D'}$, where $D'$ is the number of latent features. 
First, we investigate the effect of the feature dimensionality in our factorized spatio-temporal graph attention layer.
In this initial setting, we use a single graph attention layer and 3, 5, 7, or 9 attention heads.
The results are shown in Table ~\ref{tab:ablations} for CLEVRER and in Table~\ref{tab:ablationsAG} for Action Genome. 
Factorized attention enables us to utilize spatial and temporal information through separate channels. 
Using 256 latent dimensions works best for CLEVRER and increasing it further results in a small but consistent performance decrease. 
For Action Genome, the number of latent dimensions has less of an impact on the performance, as metric scores are rather consistent across all settings. In further experiments, we keep 256 dimensions for CLEVRER and 512 for Action Genome in the graph attention layer.

\boldparagraph{Attention heads.}
Having multiple attention heads, \ie attention mechanisms, stabilizes the learning process in attention-based approaches \cite{attentionallyouneed, GAT}. In our second study, we investigate the effect of the number of attention heads on the performance of MTD-GNN. The results are presented in Table~\ref{tab:ablations} and Table~\ref{tab:ablationsAG} for CLEVRER and Action Genome respectively. The amount of attention heads directly impacts the collision detection performance on CLEVRER across all metrics. When using three attention heads, we obtain an F1 score of 0.458 for collision detection, which improves to 0.505 with five heads. Similar behavior is observed for spatial edge types in Action Genome, where the F1 score, AP and AUC score increase from 0.344, 0.731 and 0.876 to 0.366, 0.746 and 0.883 when increasing the amount of attention heads from three to nine. A potential reason for the effectiveness of multiple heads is that it enables learning multiple dynamics in a single layer, an important ability given our setting. However, not all tasks seem to benefit consistently across all metrics. For example, in the relative motion task, using more than three attention heads benefits the AP and AUC score while decreasing the F1 score. In Action Genome, three heads lead to best performance across all three metrics in the attention edge type, whereas with the spatial edge type, nine heads is highly preferred. In the following experiments we maintain five heads for CLEVRER and nine heads for Action Genome as overall well performing configurations.

\begin{table*}
\centering
\scriptsize
\setlength{\tabcolsep}{3pt}
\renewcommand{\arraystretch}{1.0}
\newcommand{\csp}{\hskip 2em}
\floatbox[\capbeside]{table}[0.7\columnwidth]
  {\caption{\textbf{Overview of all ablation studies on ActionGenome~\cite{ji2020action}} with the following insights: i) finding the optimal model parameters for real-life data is challenging; ii) edge types with few classes benefit from multi-task learning.}
\label{tab:ablationsAG}}
{\begin{tabular}{l ccc@{\csp} ccc@{\csp} ccc}
\toprule
 & \multicolumn{3}{c}{\textbf{Attention}} & \multicolumn{3}{c}{\textbf{Spatial}} & \multicolumn{3}{c}{\textbf{Contacting}}\\
 & F1$\uparrow$ & AP$\uparrow$ & AUC$\uparrow$ & F1$\uparrow$ & AP$\uparrow$ & AUC$\uparrow$ & F1$\uparrow$ & AP$\uparrow$ & AUC$\uparrow$ \\
\rowcolor{Gray}
\multicolumn{10}{l}{\textbf{Latent Dimensions}} \\
256 & 0.365 & 0.743 & 0.702 & \textbf{0.366} & \textbf{0.746} & \textbf{0.883} & \textbf{0.371} & \textbf{0.743} & \textbf{0.963} \\
512 & \textbf{0.367} & \textbf{0.746} & \textbf{0.706} & 0.366 & 0.745 & 0.882 & 0.350 & 0.737 & 0.962 \\
1024 & 0.364 & 0.745 & 0.705 & 0.364 & 0.745 & 0.882 & 0.362 & 0.712 & 0.951 \\
\rowcolor{Gray}
\multicolumn{10}{l}{\textbf{Attention Heads}} \\
3 & \textbf{0.367} & \textbf{0.746} & \textbf{0.706} & 0.344 & 0.731 & 0.876 & 0.364 & \textbf{0.744} & \textbf{0.963} \\
5 & 0.356 & 0.740 & 0.699 & 0.361 & 0.744 & 0.882 & 0.362 & 0.743 & 0.962 \\
7 & 0.364 & 0.745 & 0.705 & 0.364 & 0.745 & 0.882 & \textbf{0.371} & 0.743 & \textbf{0.963} \\
9 & 0.364 & 0.745 & 0.705 & \textbf{0.366} & \textbf{0.746} & \textbf{0.883} & 0.360 & 0.743 & \textbf{0.963} \\
\rowcolor{Gray}
\multicolumn{10}{l}{\textbf{Attention Layers}} \\
1 & \textbf{0.367} & \textbf{0.746} & \textbf{0.706} & \textbf{0.366} & \textbf{0.746} & \textbf{0.883} & \textbf{0.371} & 0.743 & \textbf{0.963} \\
2 & 0.230 & 0.636 & 0.595 & 0.359 & 0.740 & 0.880 & 0.364 & \textbf{0.744} & \textbf{0.963} \\
3 & 0.364 & 0.744 & 0.705 & 0.327 & 0.721 & 0.872 & 0.364 & \textbf{0.744} & \textbf{0.963} \\
\rowcolor{Gray}
\multicolumn{10}{l}{\textbf{Learning Method}} \\
Single-task & \textbf{0.367} & 0.746 & 0.706 & 0.366 & \textbf{0.746} & \textbf{0.883} & \textbf{0.371} & \textbf{0.743} & \textbf{0.963}\\
Multi-task & \textbf{0.367} & \textbf{0.747} & \textbf{0.708} & \textbf{0.392} & 0.711 & 0.802 & 0.300 & 0.607 & 0.889 \\
\bottomrule
\end{tabular}}
\end{table*}

\boldparagraph{Number of aggregations.} The architecture of MTD-GNN allows for repeated spatio-temporal feature aggregation. In theory, this allow each node's features to be informed about nodes further down the graph, \ie second-, third, n-order neighbors, thereby capturing more temporal dynamics. This ablation study investigates how many aggregations are preferred. The results are shown in Table~\ref{tab:ablations} and Table~\ref{tab:ablationsAG} for CLEVRER and Action Genome respectively. Interestingly, more than one aggregation layers is not preferred across both datasets. The performance decreases consistently across recorded metrics, \eg from an F1 score of 0.505 for one aggregation level to 0.459 and 0.485 for respectively two and three levels of aggregation in collision prediction. The same phenomenon occurs across all metrics in Action Genome for attention and spatial edge types. For contacting edge types, having more than one attention layer has a negative effect on the F1 score, while slightly enhancing the AP. Using a large number of attention layers likely causes over-smoothing in the final feature representations. We conclude that it is more important to richly model a single aggregation layer with multiple attention heads and latent attention dimensions, than to model higher-order neighbourhood relations and temporal self-relations.

\boldparagraph{Multi-relational learning.} In the fourth ablation study, we investigate the importance of multi-relational modeling. In Table~\ref{tab:ablations}, results for collision detection and relative motion prediction on CLEVRER are shown. The results demonstrate that learning multiple relations in parallel enhances the prediction performance of each individual task. For collision detection, especially, the addition of the relative motion task is important, as performance improves from an F1 score of 0.505 to 0.594. The outcomes indicate that the model is able to exploit the dynamics of the scene better when presented in multi-relational manner, where weights are optimized using multiple targets. The result also aligns with our intuition that information about relative motion provides a useful cue for collision detection and vice versa.

Table~\ref{tab:ablationsAG} shows the results for the edge types in Action Genome. Interestingly, learning in multi-task setting is preferred for some edge types, but not all. Specifically, we see a performance increase for the attention edge type with three classes, at the cost of the contacting edge type with seventeen classes. This outcome is surprising, since one could argue that attention and spatial edge types provide useful cues for contacting edge types. The outcome might indicate that when edge types have different number of classes, only the one with the least classes benefits from the multi-task setting. However, the performance increase in F1-score from 0.366 to 0.392 for the spatial edge type contradicts this, since it has double as many classess as attention edge types. We conclude that the performance gain from multi-task learning in Action Genome is dependent on the edge type to predict.

\begin{wraptable}{r}{0.5\linewidth}
\vspace{-1.2cm}
\centering
\scriptsize
\setlength{\tabcolsep}{3pt}
\renewcommand{\arraystretch}{1.0}
\newcommand{\csp}{\hskip 2em}
\begin{tabular}{l ccc@{\csp} ccc}
\toprule
 & \multicolumn{3}{c}{\hspace{-1.4em}\textbf{Collision prediction}} & \multicolumn{3}{c}{\textbf{Relative motion}}\\
 & F1$\uparrow$ & AP$\uparrow$ & AUC$\uparrow$ & F1$\uparrow$ & AP$\uparrow$ & AUC$\uparrow$\\
\rowcolor{Gray}
\multicolumn{7}{l}{\textbf{Vanilla baselines}}\\
RNN~\cite{structuralrnn} & 0.247 & 0.322 & 0.527 & 0.733 & 0.710 & 0.514 \\
LSTM~\cite{lstmgraph}    & 0.289 & 0.404 & 0.595 & 0.750 & 0.805 & 0.634 \\
TCN~\cite{tgat}          & 0.345 & 0.341 & 0.548 & \textbf{0.839} & 0.708 & 0.524 \\
\rowcolor{Gray}
\multicolumn{7}{l}{\textbf{Graph attention (GA) baselines}}\\
RNN + GA  & 0.168 & 0.410 & 0.597 & 0.835 & 0.758 & 0.591 \\
LSTM + GA & 0.283 & 0.389 & 0.604 & 0.796 & 0.783 & 0.606 \\
TCN + GA  & 0.253 & 0.389 & 0.602 & 0.739 & 0.807 & 0.637 \\
\rowcolor{Gray}
\multicolumn{7}{l}{\textbf{This paper}}\\
MTD-GNN & \textbf{0.594} & \textbf{0.607} & \textbf{0.768} & \textbf{0.839} & \textbf{0.820} & \textbf{0.688}\\
\bottomrule
\end{tabular}
\vspace{-4mm}
\caption{\textbf{Comparative evaluation on CLEVRER~\cite{clevrer}.} MTD-GNN compares favorable to the baselines, showing its effectiveness for multi-relational edge prediction in temporally-dynamic spatio-temporal graphs.}
\label{tab:sota}
\end{wraptable}

\subsection{Comparative Evaluation}\label{compev}
We compare against RNN~\cite{structuralrnn}, LSTM~\cite{lstmgraph}, and TCN~\cite{tgat} architectures, along with an attention-based variant of each baseline, which are generalized to the temporally-dynamic and multi-relational nature of our problem. To enable the baselines to cope with temporally-dynamic data, each graph is padded with nodes. The amount of padded nodes depends on the maximum number of objects per frame throughout each dataset (six/ten in CLEVRER/Action Genome). Nodes are ordered by index, hence structural information is lost. Each baseline encodes the spatio-temporal graph over the temporal domain, and the final output is used to predict the edges. In the attention-based variants, the final features in the temporal encoding are used to perform feature aggregation. Here, before the edges are predicted, the time-encoded nodes are aggregated with neighboring nodes. The resulting features are fed to an edge prediction layer which predicts the edge values per relationship type.

\begin{table*}[tb]
\centering
\scriptsize
\setlength{\tabcolsep}{3pt}
\renewcommand{\arraystretch}{1.0}
\newcommand{\csp}{\hskip 2em}
\floatbox[\capbeside]{table}[0.7\columnwidth]%
  {\caption{\textbf{Comparative evaluation on Action Genome~\cite{ji2020action}.} MTD-GNN compares favorably to the baselines across nearly all metrics and edge types, showing its effectiveness on real-world data.}
  \label{tab:sotaAG}}%
{\begin{tabular}{l ccc@{\csp} ccc@{\csp} ccc}
\toprule
 & \multicolumn{3}{c}{\textbf{Attention}} & \multicolumn{3}{c}{\textbf{Spatial}} & \multicolumn{3}{c}{\textbf{Contacting}}\\
 & F1$\uparrow$ & AP$\uparrow$ & AUC$\uparrow$ & F1$\uparrow$ & AP$\uparrow$ & AUC$\uparrow$ & F1$\uparrow$ & AP$\uparrow$ & AUC$\uparrow$ \\
\rowcolor{Gray}
\multicolumn{10}{l}{\textbf{Vanilla baselines}} \\
RNN~\cite{structuralrnn} & 0.364 & 0.744 & 0.705 & 0.387 & 0.699 & 0.775 & 0.291 & 0.575 & 0.868 \\
LSTM~\cite{lstmgraph}    & 0.365 & 0.745 & 0.700 & \textbf{0.394} & 0.714 & 0.800 & 0.316 & 0.627 & 0.897 \\
TCN~\cite{tgat}          & 0.347 & 0.730 & 0.686 & 0.378 & 0.698 & 0.786 & 0.287 & 0.603 & 0.888 \\
\rowcolor{Gray}
\multicolumn{10}{l}{\textbf{Graph attention (GA) baselines}} \\
RNN + GA & 0.364 & 0.745 & 0.705 & 0.387 & 0.695 & 0.766 & 0.292 & 0.705 & 0.879 \\
LSTM + GA & 0.364 & 0.744 & 0.705 & 0.391 & 0.703 & 0.784 & 0.298 & 0.614 & 0.893 \\
TCN + GA & 0.354 & 0.728 & 0.680 & 0.363 & 0.681 & 0.781 & 0.306 & 0.598 & 0.883 \\
\rowcolor{Gray}
\multicolumn{10}{l}{\textbf{This paper}} \\
MTD-GNN & \textbf{0.367} & \textbf{0.746} & \textbf{0.706} & 0.366 & \textbf{0.746} & \textbf{0.883} & \textbf{0.371} & \textbf{0.743} & \textbf{0.963} \\
\bottomrule
\end{tabular}}
\end{table*}

Comparisons on CLEVRER and Action Genome are reported in Table~\ref{tab:sota} and Table~\ref{tab:sotaAG} respectively. For collision detection, we outperform the baselines on all metrics. This result shows the potential of modelling the dynamic spatio-temporal nature of the graphs, which is done in MTD-GNN but not in the baselines. The baselines do not benefit from additional graph attention, possibly due to aggregation of the present entities' node features with that of necessary padded nodes. We obtain an F1 score of 0.594 with MTD-GNN for the collision task, compared to F1 scores of 0.345 for the best performing baseline, namely a vanilla TCN. We observe similar gaps for AP and AUC (0.607 versus 0.410 for RNN with graph attention as best baseline and 0.768 versus 0.604 for LSTM with graph attention as best baseline). For the relative motion task, we again obtain the highest overall performance, however with smaller gaps. Especially in F1 score, some baselines perform similar. One possible reason is that the baselines learn to predict the negative class more often, causing the model to obtain a lower false positive rate compared to a model which predicts the positive class more often.

MTD-GNN also outperforms nearly all baselines on Action Genome, however usually with smaller gaps for attention and spatial edge types. A greater difference occurs when evaluated on contacting edge type classes: MTD-GNN obtains an F1 score of 0.371, compared to 0.371 of the best performing baseline, namely a TCN with graph attention. An even greater difference occurs in AP and AUC score (0.743 and 0.963 versus 0.598 and 0.883). While the baselines struggle when the number of classes per edge type is large, such as with the ``contacting'' edge type, MTD-GNN succeeds in maintaining consistent performance throughout all metrics (\eg an F1 score of 0.371 versus 0.316 for LSTM as best baseline).

\begin{wraptable}[12]{r}{0.42\linewidth}
\vspace{-0.1cm}
\centering
\scriptsize
\setlength{\tabcolsep}{1pt}
\renewcommand{\arraystretch}{1.0}
\newcommand{\csp}{\hskip 1.2em}
\begin{tabular}{l cc@{\csp} cc}
\toprule
 & \multicolumn{2}{c}{\textbf{Image}} & \multicolumn{2}{c}{\textbf{Video}}\\
 & R@20$\uparrow$ & R@50$\uparrow$ & R@20$\uparrow$ & R@50$\uparrow$ \\
\midrule
\rowcolor{Gray}
\multicolumn{5}{l}{\textbf{With ground-truth detections}} \\
VRD~\cite{VRD} & 24.92 & 25.20 & 24.63 & 24.87 \\
Freq Prior~\cite{neuralmotifs} & 45.50 & 45.67 & 44.91 & 45.05 \\
Graph R-CNN~\cite{graphrcnn} & 23.71 & 23.91 & 23.42 & 23.60 \\
MSDN~\cite{mdsn} & 48.05 & 48.32 & 47.43 & 47.67 \\
IMP~\cite{imp} & 48.20 & 48.48 & 47.58 & 47.83 \\
RelDN~\cite{reldn} & 49.37 & 49.58 & 48.80 & 48.98 \\ 
MTD-GNN (Ours) & \textbf{50.09} & \textbf{50.09} & \textbf{49.54} & \textbf{49.54} \\ 
\rowcolor{Gray}
\multicolumn{5}{l}{\textbf{Without ground-truth detections}} \\
MTD-GNN (Ours) & 46.49 & 46.49 & 46.85 & 46.85 \\
\bottomrule
\end{tabular}
\vspace{-4mm}
\caption{Comparison with SOTA methods on ActionGenome~\cite{ji2020action}.}
\label{tab:SOTAACTIONGENOME}
\end{wraptable}

Lastly, we compare against existing methods in the predicate classification task on Action Genome \cite{ji2020action}, which expect ground truth bounding boxes and object categories to predict predicate labels. We adopt our method to this ground truth setting and report results in Table~\ref{tab:SOTAACTIONGENOME}. 
MTD-GNN outperforms all existing predicate classification methods when using ground truth boxes and object categories. It also achieves noteworthy performance when predicted boxes and categories are used, as can be seen in the last row of Table~\ref{tab:SOTAACTIONGENOME}. However, this result also displays a limitation of MTD-GNN, which is that it's performance strongly depends on \\ the accuracy of the detection backbone and the accuracy of temporal linking.

The outcomes of our experiments could indicate that padding graphs is not a viable solution to deal with temporally-dynamic scenes. Preserving and aggregating original spatial and temporal dynamics displays benefits in performance. Judging from all outcomes altogether, we conclude that our approach obtains the best performance for multi-relational edge prediction in temporally-dynamic spatio-temporal graphs.

\begin{figure*}
\centering
\includegraphics[width=1\linewidth]{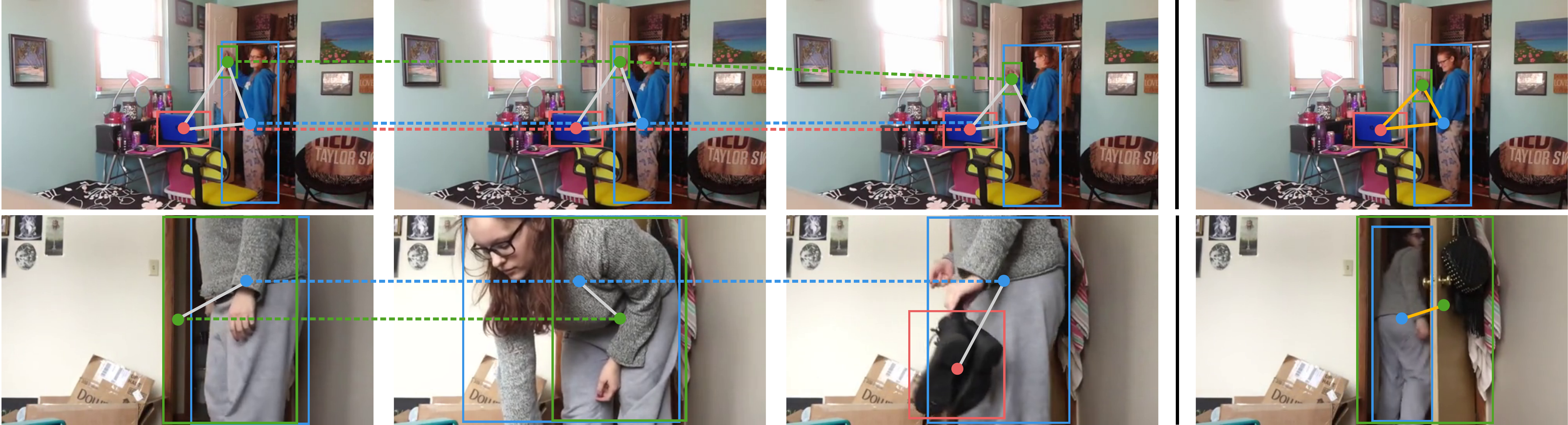}\\[-2mm]
\caption{\textbf{Qualitative analysis on ActionGenome~\cite{ji2020action}.} The top row shows a success case, where the model predicts all relations (\eg looking at laptop and holding towel) in the final frame correctly. For this, it uses the spatio-temporal graph built from the video frames before the black separation line. When objects are visible throughout the sequence, MTD-GNN benefits from temporal linking and infers relationships accurately. The bottom row shows a failure case for relations between the person and doorway. The object detector classifies the doorway as a door, which causes our method to make the wrong associations between the person and the other detected object. Hence, an inaccurate set of relationships is predicted: ``person \textit{in front of} door'' instead of ``is \textit{in} doorway''.}
\label{fig:successcases}
\end{figure*}

\subsection{Qualitative Analysis}
We perform a qualitative analysis on ActionGenome by providing success and failure cases for a multi-task MTD-GNN. Figure~\ref{fig:successcases} shows test samples where the model achieved the highest and lowest F1 score. MTD-GNN can readily account for settings with multiple visible objects, while dealing with occluded objects over time and distinct relationships of visually similar objects remain an open problem due to inaccurate temporal linking.

\newpage
\section{Conclusion}
This paper investigates how to perform edge prediction of multiple relation types simultaneously in a temporally-dynamic spatio-temporal graph network. Different from common spatio-temporal graph networks, we do not assume a single set of entities, but allow for the number of entities in the scene to vary over time. This makes our model more suitable for real-world scenarios with dynamic scenes. To address this challenging problem, we propose a factorized spatio-temporal graph attention layer. On top of this layer, we outline a multi-task optimization with an optional prioritized loss for multi-task learning. Our experiments on CLEVRER and Action Genome show that our attention-based approach can model dynamic relations in graphs, while modelling multiple relations simultaneously can be beneficial when predicting individual relations. Our approach compares favorably to approaches that can generalize to the proposed setting, as well as to state-of-the-art methods in the predicate classification task. Future research on this topic could focus on recovering from false or missed detections by the detection backbone and applying MTD-GNN to other applications, such as action forecasting.
\section{Acknowledgement}
This work has been financially supported by Mercedes-Benz, TomTom, the University of Amsterdam and the allowance of Top consortia for Knowledge and Innovation (TKIs) from the Netherlands Ministry of Economic Affairs and Climate Policy. We furthermore thank Theo Gevers, Sezer Karaoglu, Martin Oswald, Yu Wang, Ysbrand Galama and Georgi Dikov for their valuable input when writing this paper.

\bibliography{egbib}
\end{document}